\documentclass[letterpaper]{article} 
\usepackage{aaai24}  
\usepackage{times}  
\usepackage{helvet}  
\usepackage{courier}  
\usepackage[hyphens]{url}  
\usepackage{graphicx} 
\usepackage{natbib}  
\usepackage{caption} 
\usepackage{algorithm}
\usepackage{algorithmic}

\usepackage{amsmath}
\usepackage{amssymb}
\usepackage{booktabs}
\usepackage{mathrsfs}
\usepackage{multirow}
\usepackage{color}
\usepackage{bbding}
\usepackage{threeparttable}
\usepackage{xspace}
\usepackage{tabularx}
\usepackage{dcolumn}
\usepackage{array}

\newcommand{\PreserveBackslash}[1]{\let \temp =\\#1 \let \\ = \temp}
\newcolumntype{C}[1]{>{\PreserveBackslash\centering}p{#1}}

\usepackage{newfloat}
\usepackage{listings}
\DeclareCaptionStyle{ruled}{labelfont=normalfont,labelsep=colon,strut=off} 
\floatstyle{ruled}
\newfloat{listing}{tb}{lst}{}
\floatname{listing}{Listing}
\newcommand*\samethanks[1][\value{footnote}]{\footnotemark[#1]}

\title{Trend-Aware Supervision: On Learning Invariance for Semi-Supervised Facial Action Unit Intensity Estimation}
\author {
    Yingjie Chen\textsuperscript{\rm 1}\thanks{Equal Contribution.},
    Jiarui Zhang\textsuperscript{\rm 1}\samethanks,
    Tao Wang\textsuperscript{\rm 1}\thanks{Corresponding author.},
    Yun Liang\textsuperscript{\rm 2}
}
\affiliations {
    \textsuperscript{\rm 1} School of Computer Science, Peking University, Beijing China \\
    \textsuperscript{\rm 2} School of Integrated Circuits, Peking University, Beijing China \\
    chenyingjie@pku.edu.cn, zjr954@pku.edu.cn, wangtao@pku.edu.cn, ericlyun@pku.edu.cn
}

\makeatletter
\DeclareRobustCommand\onedot{\futurelet\@let@token\@onedot}
\def\@onedot{\ifx\@let@token.\else.\null\fi\xspace}
\def\ie{\emph{i.e}\onedot}
\def\eg{\emph{e.g}\onedot}
\def\Eg{\emph{E.g}\onedot}

\def\etc{\emph{etc}\onedot}
\makeatother

\usepackage{bibentry}

\begin{document}

\maketitle

\begin{abstract}
With the increasing need for facial behavior analysis, semi-supervised AU intensity estimation using only keyframe annotations has emerged as a practical and effective solution to relieve the burden of annotation.
However, the lack of annotations makes the spurious correlation problem caused by AU co-occurrences and subject variation much more prominent, leading to non-robust intensity estimation that is entangled among AUs and biased among subjects.
We observe that trend information inherent in keyframe annotations could act as extra supervision and raising the awareness of AU-specific facial appearance changing trends during training is the key to learning invariant AU-specific features.
To this end, we propose \textbf{T}rend-\textbf{A}ware \textbf{S}upervision (TAS), which pursues three kinds of trend awareness, including intra-trend ranking awareness, intra-trend speed awareness, and inter-trend subject awareness.
TAS alleviates the spurious correlation problem by raising trend awareness during training to learn AU-specific features that represent the corresponding facial appearance changes, to achieve intensity estimation invariance.
Experiments conducted on two commonly used AU benchmark datasets, BP4D and DISFA, show the effectiveness of each kind of awareness. And under trend-aware supervision, the performance can be improved without extra computational or storage costs during inference.
\end{abstract}

\section{Introduction}

\begin{figure}[!t]
    \centering
    \includegraphics[width=0.47\textwidth]{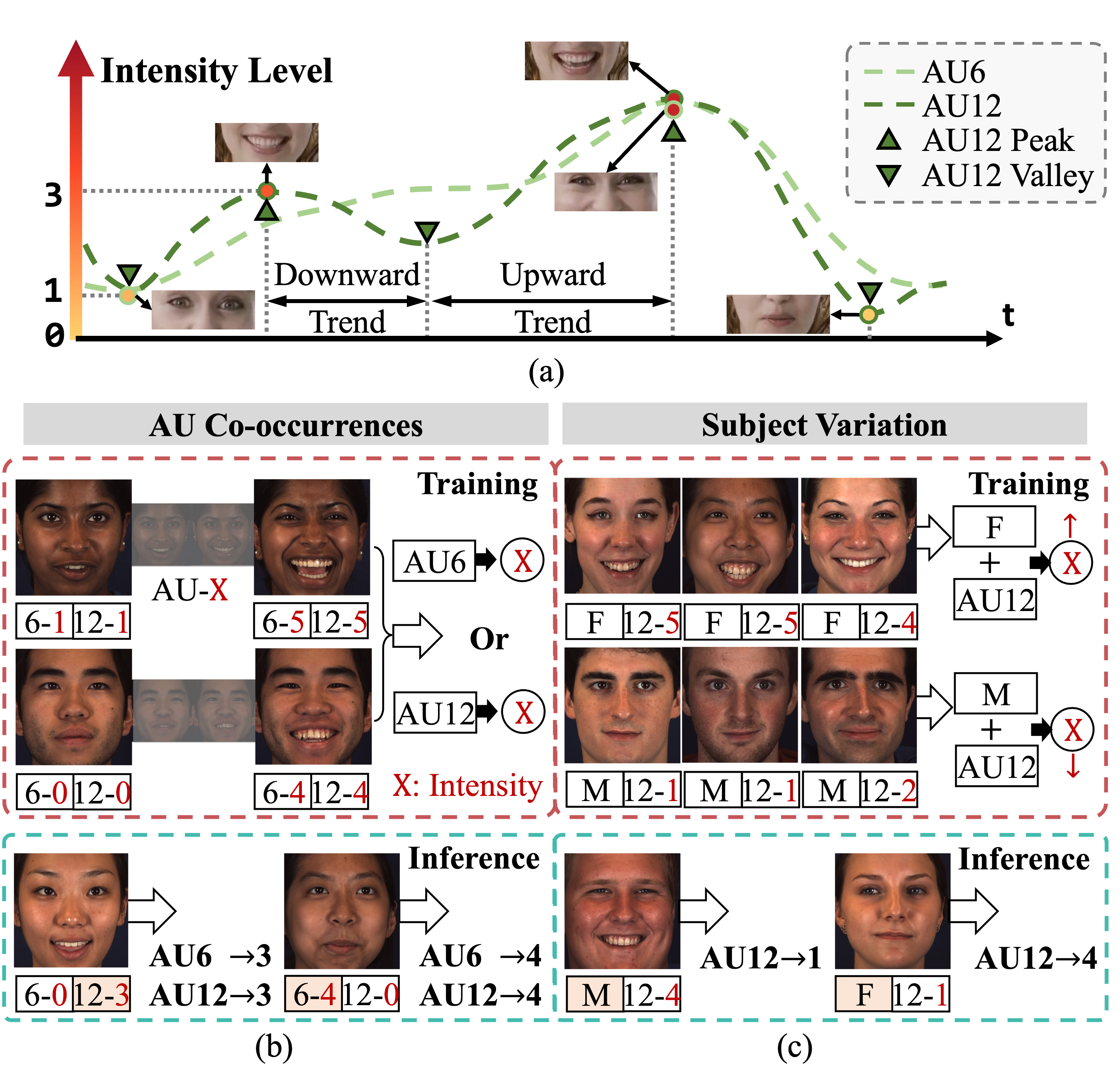}
    \caption{Motivation. (a) Illustration of keyframes and local trends. (b) Spurious correlation caused by AU co-occurrences. The limited annotations may lead the model to learn only AU features of one of the two AUs instead of both, resulting in non-robust intensity estimation entangled between them. (c) Spurious correlation caused by subject variation. Subject variation magnified by the lack of annotations may lead the model to learn non-causal gender features instead of AU features for estimating the intensity of AU12 (orange denotes the dominant features for estimation).
    }
    \label{fig:motivation}
\end{figure}

Facial behavior analysis is a fundamental task in the field of affective computing with wide application scenarios such as driver fatigue monitoring in driver assistance systems and pain estimation in medical treatment. According to FACS~\cite{friesen1978facial}, facial action units (AU) are defined as subtle facial muscle movements, and almost all facial behaviors can be expressed as certain combinations of AUs. To describe more fine-grained facial behavior, not only the states of AUs (activated or not) but also their intensity values are required. Therefore, AU intensity estimation has become a popular solution for facial behavior analysis. In FACS, the intensity values of AU are quantified into six discrete ordinal intensity levels, including neutral, trace, slight, pronounced, extreme, and maximum (from 0 to 5). Automatic AU intensity estimation aims to learn a model using FACS-quantified annotations to estimate the intensity values of multiple AUs for a facial image.

The training data requires manual labeling by FACS-certificated experts, which is time-consuming and expensive. Therefore, training AU intensity estimation models using limited annotations, referred to as semi-supervised AU intensity estimation, has attracted increasing attention.
To annotate the intensity of each AU for each frame in a video sequence, annotators usually skim through the whole video quickly and mark the frames containing a local maximum or minimum (called peak or valley) intensity value of one AU first, \ie, \textbf{keyframes} as marked in Fig.~\ref{fig:motivation} (a). Then, for each AU, with every two adjacent keyframes of this AU as references, they annotate frames between them. Considering the process, using only keyframe annotations for training has emerged as a practical and effective task setting, employed by previous works such as~\cite{zhang2018weakly,zhang2019context}. 

Semi-supervised AU intensity estimation is challenging due to the prominent spurious correlation problem~\cite{geirhos2018imagenet,hu2022improving}, \ie, non-causal correlations learned from the limited training annotations.
The trained model is supposed to only exploit AU-specific features representing the corresponding facial appearance changes, which is much harder to achieve with non-negligible spurious correlation.
\textbf{First}, the model aims to estimate the intensity values for multiple AUs simultaneously, and thus AU co-occurrences may lead to spurious correlation. 
As in Fig.~\ref{fig:motivation} (b), the model is supposed to capture both facial appearance changes from the cheek region for AU6 and lip corner region for AU12. However, their co-occurrence patterns (\eg, both AUs with the same intensity) that frequently occur in keyframe annotations may mislead the model into learning features of only one of them~\cite{shi2022robust}, \eg, using features from the lip corner region to represent both AUs and ignoring those from the cheek region. \textbf{Second}, as in Fig.~\ref{fig:motivation} (c), the lack of annotations magnifies the spurious correlation caused by subject variation. \Eg, suppose the annotated frames are mostly women with AU12 at high-intensity levels and men with AU12 at low-intensity levels. In that case, the model may be misled to use non-causal gender features instead of AU features to represent the intensity of AU12. 

We argue that the spurious correlation problem hinders the achievement of intensity estimation invariance under semi-supervised settings mainly because features of each AU are entangled with those of other AUs or subject-related features, and the lack of annotations makes it much harder to learn AU-specific features that represent the corresponding facial appearance changes.
We observe that there is rich trend information inherent in keyframe annotations, which could act as causal knowledge for capturing AU-specific facial appearance changes.
Using such trend information as extra supervision for unannotated frames to raise awareness of facial appearance changing trends during training, is the key to learning invariant AU-specific features.
To this end, we propose Trend-Aware Supervision (TAS) to make full use of the inherent trend information for pursuing both intra-trend and inter-trend awareness for AU feature learning, including intra-trend ranking awareness, intra-trend speed awareness, and inter-trend subject awareness.

Specifically, for each AU, every two adjacent keyframes divide a whole video sequence into several local trends, as shown in Fig.~\ref{fig:motivation} (a), and facial appearance corresponding to the specific AU changes monotonically in the local trend.
To alleviate the spurious correlation caused by AU co-occurrences, \emph{intra-trend ranking awareness} is pursued to distinguish the specific AU from those with significantly different trends, \ie, AUs corresponding to facial appearance that does not change monotonically.
And for the rest of AUs with similar monotonic facial appearance changing trends, \emph{intra-trend speed awareness} further distinguishes among them by encouraging the changing speed of AU features and that of the corresponding facial appearance as close as possible.
To alleviate the spurious correlation caused by subject variation, \emph{inter-trend subject awareness} strips subject information by encouraging AU features of the same class with the same intensity label but from different subjects to be as similar as possible.
Note that all kinds of awareness supervise on AU features rather than directly on intensity values, to relieve the burden of regression layers by learning invariant AU-specific features.
Combining all three awareness, here comes our trend-aware supervision scheme, which encourages the model to learn invariant correlations instead of spurious ones to achieve intensity estimation invariance.

Our main contributions are threefold. 
\begin{itemize}
    \item We inspect the keyframe-based semi-supervised AU intensity estimation problem and first identify the spurious correlation problem as the main challenge for achieving intensity estimation invariance.
    \item We observe the rich trend information inherent in keyframe annotations and propose Trend-Aware Supervision to raise intra-trend and inter-trend awareness during training to learn invariant AU-specific features.
    \item Experimental results show that without adding extra computational or storage costs in the inference stage, the performance of our model is improved and even exceeds several fully supervised ones.
\end{itemize}

\section{Related Work}
\subsection{Supervised AU Intensity Estimation}
To pursue a more precise way of describing facial behaviors, AU intensity estimation task has gradually emerged as a promising solution~\cite{rudovic2014context,mohammadi2017adaptive,sanchez2018joint,wang2020facial,song2021dynamic}. 
Works such as~\cite{li2015measuring,walecki2017deep,walecki2017copula} take advantage of Probabilistic Graphical Models to model the latent dependencies among AUs.
\cite{wang2018facial} proposed HBN that employs a hybrid Bayesian Network to estimate AU intensity values by capturing the global dependencies among AUs.

Works such as~\cite{kaltwang2015doubly,linh2017deepcoder,fan2020joint,fan2020facial,fan2021g2rl} make efforts in learning AU-specific features by injecting prior knowledge into models.
For using location information, \cite{fan2020joint} proposed a joint AU intensity prediction and localization method that works directly on the whole input image.
And works such as~\cite{fan2020facial,fan2021g2rl} transfer AU intensity estimation task into a facial landmark heatmap regression problem. Instead of introducing explicit multi-task or joint learning, our model learns better AU-specific features implicitly via a trend-aware training process.


\subsection{Semi-Supervised AU Intensity Estimation}
In the face of the overwhelming pressure of manually annotating a large amount of training data, semi-supervised AU intensity estimation has become a practical and effective solution~\cite{zhao2016facial,zhang2018bilateral,zhang2018weakly,wang2019deep,zhang2019context,sanchez2020semi}.

Due to the lack of annotations, prior knowledge plays an essential role in improving performance. 
OSVR~\cite{zhao2016facial} exploits the ordinal information among different frames and intensity labels of selected frames, but it requires pre-extracted features as input and training models separately for each AU.
BORMIR~\cite{zhang2018bilateral} learns an extra relevant value for each frame and not only constraints ordinal relevance but also constraints relevance smoothness and intensity smoothness via a regularization term. 
KBSS~\cite{zhang2018weakly} uses segments containing only four sampled frames with extra neutral frames for training, resulting in insufficient use of ordinal information.
To capture fine-grained AU features, \cite{zhang2019context} made efforts in learning context-aware features and proposed an LSTM-based method with learnable AU-related context for AU-specific feature learning. 
We argue that constraining feature or intensity value smoothness via a regularization term as previous works may lead to the over-smooth problem. In contrast, we propose intra-trend speed awareness to achieve smooth and stable estimation without breaking away from the corresponding facial appearance changing trend.

Despite their success, previous works failed in solving the spurious correlation problem, leading to non-robust estimation. In this paper, we propose Trend-Aware Supervision to achieve intensity estimation invariance.

\section{Approach}
\begin{figure*}[!t]
    \centering
    \includegraphics[width=0.96\textwidth]{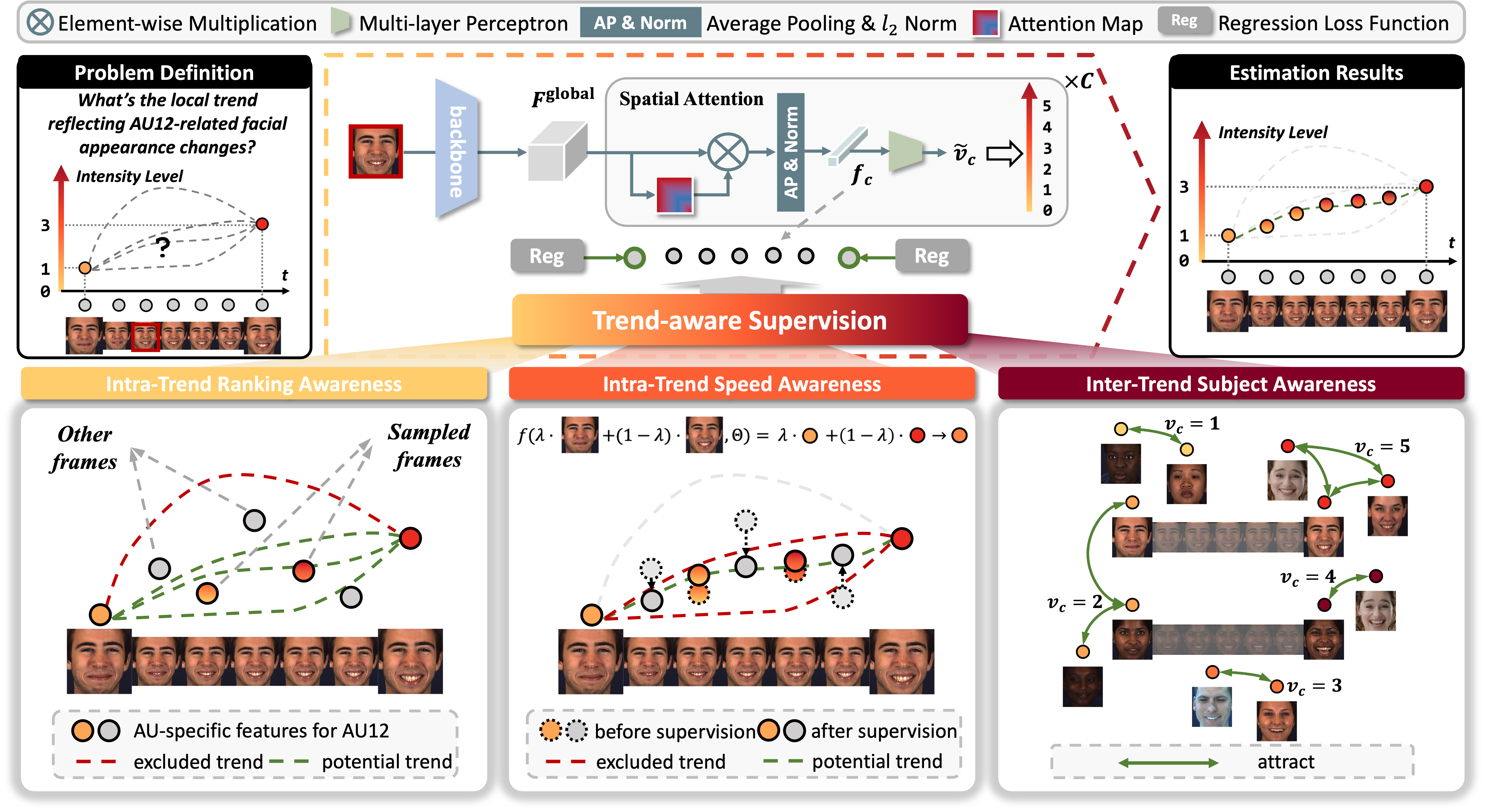}
    \caption{Overview. Our method takes a batch of segments as input and each image (taken the one with red box as an example) is first fed into a backbone network for global feature extraction. Then, $C$ separate spatial attention layers are applied to $F_{\rm{global}}$ for AU feature extraction. After that, each AU feature $f_c$ is fed into an MLP to estimate the intensity value $\Tilde{v}_c$ for the $c^{\rm th}$ AU class. During training, trend-aware supervision is applied to AU features $\{f_c^t\}^{T}_{t=1}$ in each segment. And regression loss function is applied to the estimated intensity results of the two annotated keyframes only, \ie, the first and the last ones in each segment.}
    \label{fig:overview}
\end{figure*}

\subsection{Problem Formulation}
AU intensity estimation aims to train a model $f(I, \Theta)$ that estimates the intensity values of $C$ AU classes $\Tilde{\mathbf{v}}=\{\Tilde{v}_c\in \mathbb{R}\}^{C}_{c=1}$ for a given facial image simultaneously.
Under the keyframe-based setting, video sequences with only keyframe annotations are used as training data, \ie, for each frame in a video sequence, FACS-quantified intensity label $v_c\in\mathbb{N}$ is given for the current frame only if the intensity of the $c^{\rm th}$ AU reaches peak or valley.
The goal is to find an optimal $\Theta$ that makes the estimated intensity $\Tilde{\mathbf{v}}$ as close to the real trend of AU-specific facial appearance changes as possible, based on FACS-quantified keyframe annotations.

To train our AU intensity estimation model, for each AU class, we split those video sequences into segments based on the locations of keyframes, the same as~\cite{zhang2019context}. In each segment for a specific AU, the intensity of this AU evolves either from peak to valley or from valley to peak. For each segment, only the linearly sampled $T$ frames are used.
In this way, the training data is organized into segments $\{\mathbf{I}_i, v_{i, c}^1, v_{i, c}^T\}^{N}_{i=1}$, where $N$ denotes the number of segments, $c\in \{1, 2, \ldots, C\}$ denotes the $c^{\rm th}$ AU out of the total $C$ AU classes, and $\mathbf{I} = \{I^t\in \mathbb{R}^{H\times W\times 3}\}^{T}_{t=1}$ is the $T$ images contained in each segment.
Note that only the intensity label of the $c^{\rm th}$ AU in the first and the last images are given, \ie, $\{v_c^1, v_c^T\}$. For convenience, the intensity labels are uniformly normalized from range $[0, 5]$ to range $[0, 1]$.

\subsection{Overview}
As shown in Fig.~\ref{fig:overview}, our goal is to learn an optimal $\Theta$ for an image-based AU intensity estimation model $f(I, \Theta)$. 
In the training stage, a batch of pre-processed segments is taken as input. Each image is fed into a backbone network to extract global feature $F^{\rm global}\in \mathbb{R}^{d\times H_{\rm g}\times W_{\rm g}}$. Then, to obtain AU features $F^{\rm au}=\{f_c\in \mathbb{R}^{d}\}^{C}_{c=1}$, $C$ separate spatial attention layers~\cite{zhao2019pyramid} are applied to the extracted $F^{\rm global}$, and through the corresponding $c^{\rm th}$ branch, AU feature $f_c$ is obtained after an average pooling operation and an $l_2$ normalization operation. After that, AU feature $f_c$ is fed into the corresponding MLP consisting of two linear layers to estimate the intensity value $\Tilde{v}_c\in \mathbb{R}$ for the $c^{\rm th}$ AU class, which is clipped to range $[0, 1]$. In this way, the final intensity estimation results $\Tilde{\mathbf{v}}=\{\Tilde{v}_c\}^{C}_{c=1}$ are obtained.

During training, a commonly used regression loss function acts as direct supervision for the two annotated keyframes in each segment, for which we employ Mean-squared Error (MSE) loss function as shown in Eq.~\ref{eq:regression}. Besides, we apply the proposed trend-aware supervision to the AU features $F^{\rm au}$ extracted from each image, to make up for the lack of annotations by raising trend awareness for invariant AU-specific feature learning.

\begin{equation}
    \mathcal{L}_{\rm reg} = \sum^{N}_{i=1}{\left((v_{i, c}^1-\Tilde{v}_{i, c}^1)^2 + (v_{i, c}^T-\Tilde{v}_{i, c}^T)^2\right)}.
\label{eq:regression}
\end{equation}

In the inference stage, \emph{note that segment input is not required anymore}. The obtained model $f(I, \Theta)$ takes one facial image as input and outputs estimation result $\Tilde{\mathbf{v}}$ directly.

\subsection{Trend-Aware Supervision}
We propose trend-aware supervision to achieve intensity estimation invariance, which pursues three kinds of awareness. Different from previous works~\cite{zhang2018bilateral,zhang2019joint} that directly constrain smoothness or ordering of intensity values, we first focus on learning invariant AU-specific features for more robust intensity estimation.

\subsubsection{Intra-Trend Ranking Awareness}
As aforementioned, keyframe annotations of each AU divide a whole video sequence into several local trends, and between pairs of adjacent keyframes, facial appearance corresponds to the specific AU changes monotonically, which is essential supervision for distinguishing the AU from others that correspond to non-monotonic changing trends of AU-specific facial appearance.
Instead of focusing on constraining intensity values~\cite{zhang2018bilateral,zhang2019joint}, we first focus on constraining AU-specific features.
We argue that constraining intensity values directly will make the MLP layers do a lot of heavy lifting, leading to less specificity in the extracted AU features.
Therefore, instead of directly forcing the estimated results to be monotonic, we enforce the changing amplitudes of AU features relative to the first frame to be monotonic, to make it correspond to the changing amplitudes of AU-specific facial appearance, which benefits the model to capture facial appearance features from regions related to the specific AU, \ie, making AU features more sensitive to the intensity changes of the corresponding facial appearance.

To measure the changing amplitudes of AU features, we compute the Euclidean distance between the AU feature of each frame and that of the first frame, $\Delta \mathbf{f}_c = \{\Delta f_c^t={\Vert f_c^t - f_c^1\Vert}_2\}^{T}_{t=1}$, and enforce a monotonic increasing trend among them, \ie, $\Delta f_c^1 \leq \Delta f_c^2, \dots, \leq \Delta f_c^T$. Note that whether the trend is upward or downward, the corresponding facial appearance changing amplitudes relative to the first frame keep getting larger. In this way, the loss function for intra-trend ranking awareness is defined as in Eq.~\ref{eq:monotonic}.

\begin{equation}
    \mathcal{L}_{\rm rank}=\sum^{N}_{i=1}\sum^{T-1}_{t=1}{\operatorname{Max}(\mathbf{0}, \mathbf{A}\Delta\mathbf{f}_{i,c})},
\label{eq:monotonic}
\end{equation}

where $\mathbf{A}_{T\times T}$ is for computing the difference of first order of $\Delta\mathbf{f}_{i,c}$ by setting $\{a_{t, t}\}^{T}_{t=1}$ to $1$, $\{a_{t, t+1}\}^{T-1}_{t=1}$ to $-1$, and others to $0$. Note that for each segment, only AU features of the annotated AU class are constrained.

\begin{table*}[!t]
\setlength\tabcolsep{3.2pt}
\small
\centering
\begin{tabular}{cc|ccccc|c|cccccccccccc|c}
    \toprule
    \multicolumn{2}{c|}{Dataset} & \multicolumn{6}{c}{BP4D} & 
    \multicolumn{13}{|c}{DISFA}\\
    \midrule
     Metric & Method & AU6 & 10 & 12 & 14 & 17 & \textbf{Avg.} & AU1 & 2 & 4 & 5 & 6 & 9 & 12 & 15 & 17 & 20 & 25 & 26 & \textbf{Avg.} \\
    \midrule
    \multirow{8}{*}{ICC$\uparrow$} 
    & Ladder~\nocite{rasmus2015semi} & .67 & .62 & .79 & .07 & .44 & .52 & -.01 & .06 & .04 & .03 & .46 & .09 & .60 & -.02 & .01 & .00 & .58 & .37 & .18 \\
    & OSVR~\nocite{zhao2016facial} & .65 & .58 & .78 & .27 & .45 & .54 & .21 & .04 & .25 & .15 & .23 & .15 & .31 & .12 & .07 & .09 & .62 & .09 & .19 \\
    & BORMIR~\nocite{zhang2018bilateral} & .73 & .68 & .86 & .37 & .47 & .62 & .20 & .25 & .30 & .17 & .39 & .18 & .58 & .16 & .23 & .09 & .71 & .15 & .28 \\
    & KJRE~\nocite{zhang2019joint} ($6\%$) & .71 & .61 & \underline{.87} & .39 & .42 & .60 & .27 & \underline{.35} & .25 & .33 & .51 & .31 & .67 & .14 & .17 & \underline{.20} & .74 & .25 & .35 \\
    & KBSS~\nocite{zhang2018weakly} & .76 & .73 & .84 & \underline{.45} & .45 & .65 & .14 & .12 & .48 & .17 & .43 & .35 & .71 & .15 & .25 & .09 & .78 & .54 & .35 \\
    & CFLF~\nocite{zhang2019context} & \underline{.77} & .70 & .83 & .41 & \underline{.60} & .66 & .26 & .19 & .46 & \underline{.35} & .52 & \underline{.36} & .71 & \underline{.18} & \underline{.34} & \textbf{.21} & .81 & .51 & .41 \\
    & RandCon~\nocite{sanchez2020semi} & \underline{.77} & \underline{.75} & .86 & \textbf{.48} & .55 & \underline{.68} & \underline{.33} & .33 & \underline{.65} & -.02 & \underline{.60} & .34 & \underline{.78} & .18 & .24 & .08 & \underline{.88} & \underline{.58} & \underline{.41} \\
    & Ours & \textbf{.78} & \textbf{.77} & \textbf{.88} & \textbf{.48} & \textbf{.62} & \textbf{.71} & \textbf{.47} & \textbf{.57} & \textbf{.70} & \textbf{.38} & \textbf{.62} & \textbf{.37} & \textbf{.80} & \textbf{.25} & \textbf{.41} & .14 & \textbf{.92} & \textbf{.59} & \textbf{.52} \\
    \midrule
    \multirow{8}{*}{MAE$\downarrow$}
    & Ladder~\nocite{rasmus2015semi} & .69 & .84 & \underline{.60} & 1.20 & \underline{.64} & .79 & .65 & .34 & 1.26 & \textbf{.11} & \textbf{.28} & .33 & \textbf{.35} & \underline{.19} & \underline{.30} & \textbf{.15} & .76 & .39 & .43 \\
    & OSVR~\nocite{zhao2016facial} & 1.02 & 1.13 & .95 & 1.35 & .93 & 1.08 & 1.65 & 1.87 & 2.94 & 1.38 & 1.56 & 1.69 & 1.64 & 1.10 & 1.61 & 1.37 & 1.33 & 1.79 & 1.66 \\
    & BORMIR~\nocite{zhang2018bilateral} & .85 & .90 & .68 & 1.05 & .79 & .85 & .88 & .78 & 1.24 & .59 & .77 & .78 & .76 & .56 & .72 & .63 & .90 & .88 & .79 \\
    & KJRE~\nocite{zhang2019joint} ($6\%$) & .82 & .95 & .64 & 1.08 & .85 & .87 & 1.02 & .92 & 1.86 & .70 & .79 & .87 & .77 & .60 & .80 & .72 & .96 & .94 & .91 \\
    & KBSS~\nocite{zhang2018weakly} & .74 & \underline{.77} & .69 & \underline{.99} & .90 & .82 & .53 & .49 & .82 & .24 & .39 & .38 & .43 & .32 & .50 & .36 & .61 & .44 & .46 \\
    & CFLF~\nocite{zhang2019context} & \underline{.62} & .83 & .62 & 1.00 & \textbf{.63} & \underline{.74} & \textbf{.33} & \textbf{.28} & \textbf{.61} & \underline{.13} & .35 & \underline{.28} & .43 & \textbf{.18} & \textbf{.29} & \underline{.16} & .53 & .40 & \textbf{.33} \\
    & RandCon~\nocite{sanchez2020semi} & .65 & .91 & .83 & \textbf{.98} & \textbf{.63} & .80 & \underline{.43} & \underline{.36} & \underline{.65} & .19 & .38 & .37 & .46 & .25 & .38 & .21 & \underline{.45} & \underline{.39} & .38 \\
    & Ours & \textbf{.53} & \textbf{.67} & \textbf{.51} & \underline{.99} & .73 & \textbf{.69} & .52 & \underline{.36} & \textbf{.61} & .19 & \underline{.33} & \textbf{.27} & \underline{.39} & .23 & .46 & .27 & \textbf{.33} & \textbf{.35} & \underline{.36} \\
    \bottomrule
\end{tabular}
\caption{Comparison with the state-of-the-art semi-supervised methods. The best and second results are indicated using bold and underline, respectively. Note that KJRE uses $6\%$ of the annotations while others use $2\%$.}
\label{tab:sota_semi}
\end{table*}

\subsubsection{Intra-Trend Speed Awareness}
Intra-trend ranking awareness is not enough to distinguish the AU from those that also correspond to a monotonic changing trend of AU-specific facial appearance.
Therefore, intra-trend speed awareness is pursued to raise the awareness of facial appearance changing speed by encouraging the model to behave linearly in between frames in a local trend. 
In other words, if the AU-specific facial appearance changes rapidly or slowly in a local trend, the changing amplitudes of AU features relative to the first frame are also supposed to change rapidly or slowly.
The lack of annotations increases the difficulty of raising such speed awareness since there is no direct supervision for frames between adjacent keyframes. We make up for that by modeling the vicinity relation across images in one local trend. 
Given \emph{mixup} operation~\cite{zhang2018mixup}: $\mathcal{M}_{\lambda}(x_i, x_j)=\lambda \cdot x_i + (1-\lambda)\cdot x_j$, we sample virtual image-target pairs from the \emph{mixup} vicinal distribution and encourage the model to behave linearly in-between the sampled images in each segment, \ie, encouraging the model to estimate an intermediate intensity $\mathcal{M}_{\lambda}(f(I^i, \Theta), f(I^j, \Theta))$ for input $\mathcal{M}_{\lambda}(I^i, I^j)$ ($I^i$ and $I^j$ are from the same segment).

The benefits lie in two folds: 1) for limited annotations, the augmented virtual image-target pairs encourage the model to learn decision boundaries that transit linearly between quantified intensity labels, providing a smoother estimation.
2) for AU-specific feature learning, since $\mathcal{M}$ is only applied between images in a segment of one subject, the features relevant to subject identity are almost unchanged and the features relevant to AU-specific facial appearance changes are highlighted. Thus, the model benefits from learning AU-specific features coupling with intensity changes, promoting stable estimation reflecting the real AU-specific facial appearance changes.

In each iteration, for an input segment with the estimated intensity results $\{\Tilde{\mathbf{v}}^t=f(I^t, \Theta)\}^{T}_{t=1}$, we shuffle images to construct virtual image-target pairs, $(\mathcal{M}_{\lambda}(I^i,I^j), \mathcal{M}_{\lambda}(\Tilde{\mathbf{v}}^i,\Tilde{\mathbf{v}}^j))$, and define the loss function as:

\begin{equation}
    \mathcal{L}_{\rm spd} = \sum^{N}_{n=1}\sum_{i, j}{\Vert{f\left(\mathcal{M}_{\lambda}(I^i,I^j),\Theta\right)-\mathcal{M}_{\lambda}(\Tilde{\mathbf{v}}^i,\Tilde{\mathbf{v}}^j))}\Vert^2_2},
\label{eq:speed}
\end{equation}

where $\lambda\sim Beta(\alpha, \alpha)$ and $\alpha$ is a hyper-parameter, and for $i, j$, frame indexes are shuffled and matched with the original ones to construct $T$ pairs for each segment.

\subsubsection{Inter-Trend Subject Awareness}
By raising the above two awareness during training, we focus on promoting the coupling between intra-trend AU-specific facial appearance changes and AU feature changes, making the learned AU-specific features reflect the real trend of the corresponding facial appearance changes.
However, due to spurious correlation caused by subject variation, the model has no incentive to learn invariant AU-specific features for estimation as learning some spurious features (such as gender, race, \etc) suffices to estimate target AU intensity. Therefore, AU features of the same class with the same intensity labels still vary among local trends from different subjects, which makes the model hard to generalize well on unseen subjects. To tackle the problem, AU-specific features that are invariant among subjects are required to be learned.

To learn invariant AU-specific features that are not entangled with subject-related ones, it is important to strip subject-related features by aligning AU features of one class from different subjects but with the same intensity label, \ie, only retaining AU-specific facial appearance features invariant among subjects for achieving intensity estimation invariance.
To this end, we pursue inter-trend subject awareness during training for AU-specific feature learning. The object function is clear that AU features of the same class with the same intensity label are supposed to be similar, no matter from which subjects they are extracted. And considering that only keyframes are with reliable annotations, the loss function for inter-trend subject awareness is only applied to keyframes of different segments, as shown in Eq.~\ref{eq:subject}:
\begin{equation}
    \mathcal{L}_{\rm sub} = \frac{1}{C}\sum^{C}_{c=1}\mathbb{E}_{v_c}\left[(1 - f_{i,c}^t\cdot f_{j,c}^k)\right],
\label{eq:subject}
\end{equation}

where $t, k\in \{1, T\}$ are annotated keyframes in segment $i$ and $j$ with the same intensity label $v_c$.

The overall loss function for training is as follows:

\begin{equation}
    \mathcal{L}_{\rm all} = \mathcal{L}_{\rm reg} + \lambda_{\rm rank}\mathcal{L}_{\rm rank} + \lambda_{\rm spd}\mathcal{L}_{\rm spd} + \lambda_{\rm sub}\mathcal{L}_{\rm sub},
\label{eq:overall}
\end{equation}

where $\lambda_{\rm rank}$, $\lambda_{\rm spd}$, and $\lambda_{\rm sub}$ are for balancing.

\begin{figure*}[!t]
    \centering
    \includegraphics[width=1.0\textwidth]{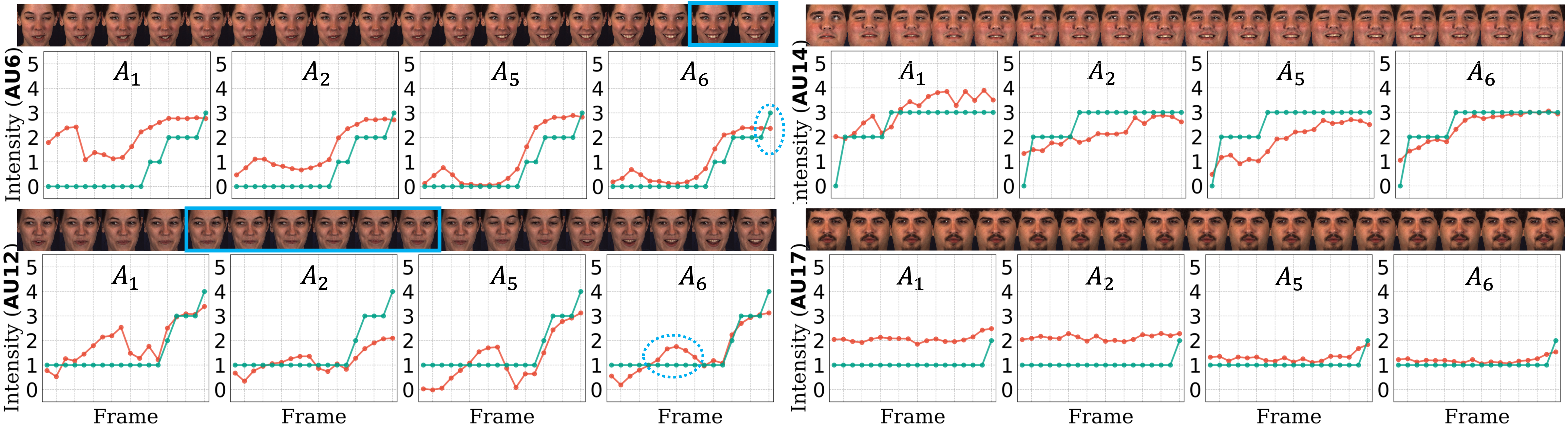}
    \caption{Case study. In each tuple, from left to right, four line charts show the intensity values estimated by model $A_1$, $A_2$, $A_5$ and $A_6$ (red line) for the given sequences on the top, respectively, and the FACS-quantified intensity labels (green line).
    }
    \label{fig:ablation_case}
\end{figure*}

\section{Experiments}
\subsection{Experimental Settings}
We use two AU benchmark datasets for evaluation, BP4D~\cite{zhang2014bp4d} and DISFA~\cite{mavadati2013disfa}.
\textbf{BP4D} for FERA 2015 challenge~\cite{valstar2015fera}, involves 23 female and 18 male subjects. Around 140,000 frames are annotated by two FACS-certified coders with quantified AU intensity labels for 5 AU classes. As the same as works~\cite{zhang2018weakly,zhang2019context}, the official split is used, in which 21 subjects are selected for training and the rest 20 subjects are for testing.
\textbf{DISFA} is a spontaneous AU dataset consisting of 26 adult subjects. Around 130,000 image frames are annotated with quantified AU intensity labels for 12 AU classes. For a fair comparison, we follow the same experimental setting mentioned in~\cite{zhang2018weakly,zhang2019context}, conducting a subject-exclusive 3-fold cross-validation.
FACS-quantified AU intensity labels of both datasets are within a discrete scale from 0 to 5.
\textbf{Only about $2\%$ annotations in BP4D and $1\%$ in DISFA} --- keyframe annotations for each AU are used for our semi-supervised setting, the same as~\cite{zhang2019context}.
Intra-class Correlation (ICC(3,1))~\cite{shrout1979intraclass} and Mean Absolute Error (MAE) are reported, the same as previous works such as~\cite{zhao2016facial,zhang2018bilateral,zhang2019context}. 

\paragraph{Implementation Details}
For each input image, Dlib~\cite{king2009dlib} is used for aligning facial images. Each image is cropped and resized to $256\times 256$.
We use ResNet34~\cite{he2016deep} without final layers as the backbone network. For each spatial attention layer, the kernel size is set to 9. $H_{\rm g}=W_{\rm g}=8, d=512$ for the extracted global feature. Each MLP for intensity estimation consists of two linear layers ($512\rightarrow 64, 64\rightarrow 1$) with LeakyReLU activation function between them, and a clipping function is applied to its output to clip values into range $[0,1]$. Empirically, $T$, $\alpha$, $\lambda_{\rm rank}$ and $\lambda_{\rm spd}$ are set to 16, 0.5, 0.1, and 0.05, respectively. $\lambda_{\rm sub}$ is set to 0.005 on BP4D and 0.05 on DISFA due to the different number of subjects.

We implement our method on Pytorch~\cite{pytorch} platform. SGD is used for optimization with weight decay of 0.0005 and learning rate of 0.005. We set batch size to 16 for both datasets. Each model is trained for 20 epochs, and early stopping strategy is adopted. All the experiments are conducted on one NVIDIA A100 Tensor Core GPU. 

\subsection{Comparison to the State-of-the-arts}

\emph{Comparison to semi-supervised methods.} First, we compare our method with top-performing semi-supervised AU intensity estimation methods, including Ladder~\cite{rasmus2015semi},
OSVR~\cite{zhao2016facial},  BORMIR~\cite{zhang2018bilateral},
KJRE~\cite{zhang2019joint},
KBSS~\cite{zhang2018weakly}, 
and CFLF~\cite{zhang2019context}, as shown in Table~\ref{tab:sota_semi}. 
In Table~\ref{tab:sota_semi}, our method achieves the highest ICC and the lowest MAE on both datasets, and both metrics outperform the compared methods by a large margin. 

\emph{Comparison to fully-supervised methods.} We also compare our method to several state-of-the-art fully-supervised methods, including HBN~\cite{wang2018facial}, Heatmap~\cite{sanchez2018joint}, 2DC~\cite{linh2017deepcoder}, 
CCNN-IT~\cite{walecki2017deep}, 
CNN~\cite{gudi2015deep}, 
RE-Net~\cite{yang2020RENet},
and SCC~\cite{fan2020facial},
All of them use full annotations for training, which is a much larger amount of annotations than that used for our setting. As in Table~\ref{tab:sota_fully}, our method achieves comparable ICC and MAE to other methods, considering the limited annotations.

\begin{table}[!t]
\centering
\begin{tabular}{c|cc|cc}
    \toprule
    Dataset & \multicolumn{2}{c}{BP4D} & 
    \multicolumn{2}{|c}{DISFA}\\
    \midrule
    Method & ICC & MAE & ICC & MAE \\
    \midrule
    HBN~\nocite{wang2018facial} & .70 & - & - & - \\
    Heatmap~\nocite{sanchez2018joint} & .68 & - & - & - \\
    2DC~\nocite{linh2017deepcoder} & .66 & - & .49 & - \\
    CCNN-IT~\nocite{walecki2017deep} & .63 & 1.26 & .38 & .66 \\
    CNN~\nocite{gudi2015deep} & .60 & .82 & .33 & .42 \\
    RE-Net~\nocite{yang2020RENet} & .64 & \underline{.65} & \textbf{.54} & \underline{.22} \\
    SCC~\nocite{fan2020facial} & \textbf{.72} & \textbf{.58} & .47 & \textbf{.20} \\
    Ours & \underline{.71} & .69 & \underline{.52} & .36 \\
    \bottomrule
\end{tabular}
\caption{Comparison with supervised methods. The best and second results are indicated using bold and underline.}
\label{tab:sota_fully}
\end{table}

\begin{table}[!b]
\setlength\tabcolsep{3.6pt}
\small
\centering
\begin{tabular}{c|c|c|c|c|cc|cc}
    \toprule
    & \multicolumn{4}{c|}{Dataset} & \multicolumn{2}{c}{BP4D} & 
    \multicolumn{2}{|c}{DISFA}\\
    \midrule
    & \multicolumn{1}{c|}{\emph{Sep}} & \multicolumn{1}{c|}{$\mathcal{L}_{\rm rank}$} & \multicolumn{1}{c|}{$\mathcal{L}_{\rm spd}$} & \multicolumn{1}{c|}{$\mathcal{L}_{\rm sub}$} & ICC & MAE & ICC & MAE \\
    \midrule
    $A_0$ &  &  &  &  & .59 & .92 & .37 & .89 \\
    $A_1$ & \Checkmark &  &  &  & .68 & .84 & .44 & .63 \\
    $A_2$ & \Checkmark & \Checkmark &  &  & .69 & .78 & .48 & .60 \\
    $A_3$ & \Checkmark &  & \Checkmark &  & .70 & .75 & .48 & .57 \\
    $A_4$ & \Checkmark &  &  & \Checkmark & .68 & .79 & .46 & .55 \\
    $A_5$ & \Checkmark & \Checkmark & \Checkmark &  & .70 & .71 & .50 & .47 \\
    $A_6$ & \Checkmark & \Checkmark & \Checkmark & \Checkmark & \textbf{.71} & \textbf{.69} & \textbf{.52} & \textbf{.36} \\
    \bottomrule
\end{tabular}
\caption{Ablation study on $\mathcal{L}_{\rm rank}$, $\mathcal{L}_{\rm spd}$, and $\mathcal{L}_{\rm sub}$. \emph{Sep} denotes separate spatial attention layers.}
\label{tab:ablation}
\end{table}


\begin{table}[!ht]
\small
\centering
\begin{tabular}{c|C{1.1cm}C{1.1cm}|C{1.1cm}c}
    \toprule
    \multicolumn{1}{c}{Dataset} & \multicolumn{2}{|c}{BP4D} & 
    \multicolumn{2}{|c}{DISFA} \\
    \midrule
    \multicolumn{1}{c|}{Method} & ICC & MAE & ICC & MAE \\
    \midrule
    \multicolumn{5}{c}{$S_1$: segments, all frames, multi AU labels} \\
    \midrule
    Baseline & .65 & .68 & .49 & .31 \\ 
    + $\mathcal{L}_{\rm all}$ & \textbf{.68} & \textbf{.63} & \textbf{.52} & \textbf{.29} \\
    \midrule
    \multicolumn{5}{C{7.8cm}}{$S_2$: segments, randomly selected 2 frames, single AU label} \\
    \midrule
    Baseline & .67 & .80 & .42 & .55 \\
    + $\mathcal{L}_{\rm all}$ & \textbf{.68} & \textbf{.73} & \textbf{.48} & \textbf{.49} \\
    \midrule
    \multicolumn{5}{C{7.8cm}}{$S_3$: randomly selected $2\%$ ($1\%$) frames, multi AU labels} \\
    \midrule
    Baseline & .67 & .67 & .44 & .27 \\
    + $\mathcal{L}_{\rm sub}$ & \textbf{.69} & \textbf{.63} & \textbf{.47} & \textbf{.25} \\
    \bottomrule
\end{tabular}
\caption{Ablation study on other semi-supervised settings.}
\label{tab:settings}
\end{table}

\begin{figure*}[!t]
    \centering
    \includegraphics[width=1.0\textwidth]{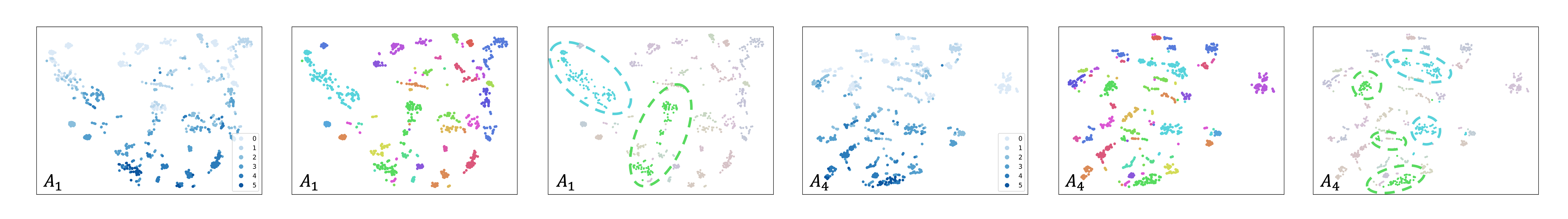}
    \caption{t-SNE visualization for AU features of AU12 on BP4D. From left to right, every three t-SNE results are colored according to FACS-quantified intensity labels (light blue to dark blue), subject identities (bright colors), and highlighted subject identities (bright blue and green). The first three are for Model $A_1$, and the last three are for Model $A_4$.
    }
    \label{fig:tsne_sub}
\end{figure*}

\begin{figure}[!ht]
    \centering
    \includegraphics[width=0.47\textwidth]{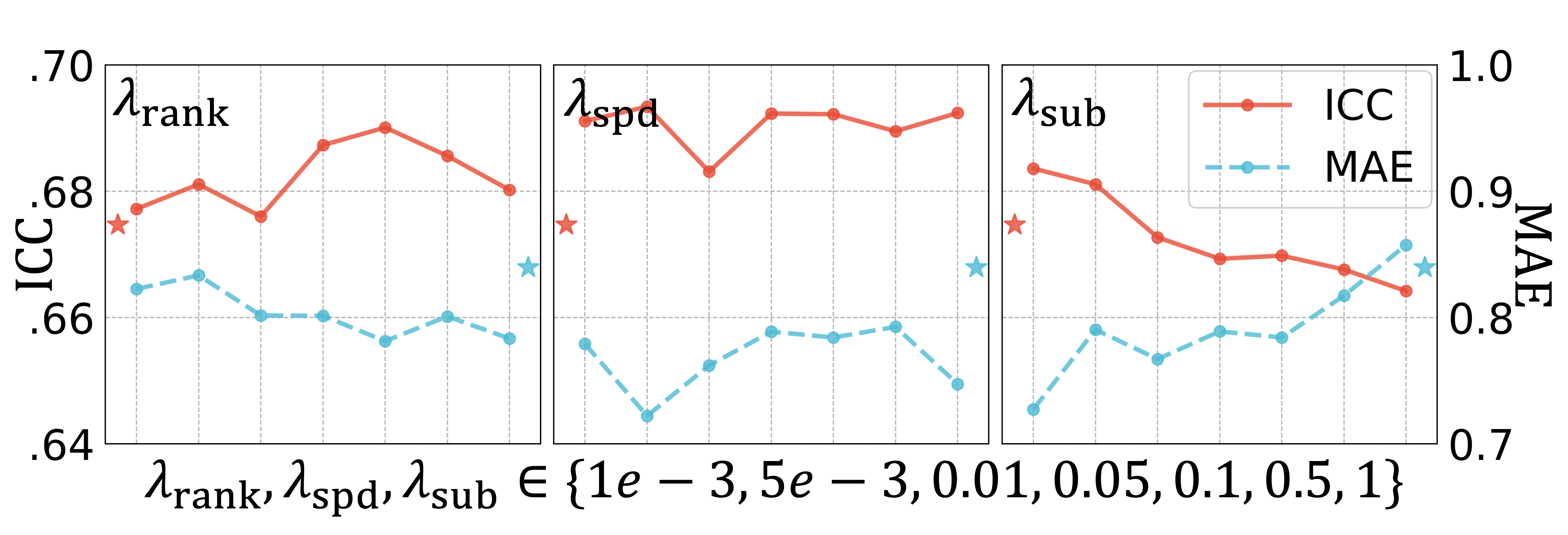}
    \caption{Empirical study for $\lambda_{\rm rank}$, $\lambda_{\rm spd}$, and $\lambda_{\rm sub}$ on BP4D. Red and blue stars show ICC and MAE of $A_1$.
    }
    \label{fig:hyper}
\end{figure}

\subsection{Ablation Study}
To validate the effectiveness of all kinds of awareness, we conduct an ablation study in Table~\ref{tab:ablation}. Model $A_0$ is vanilla ResNet34 without spatial attention layers, and other models ($A_1$ to $A_6$) are with the same network architecture as in Fig.~\ref{fig:overview} and only differ in the applied loss function. It can be observed that each awareness contributes to performance improvement, and by adding them one by one ($A_2$, $A_5$, $A_6$), Model $A_6$ achieves the best performance.

To further demonstrate the superiority of trend-aware supervision, we apply it to baseline models trained under other semi-supervised settings ($S_1, S_2, S_3$), as shown in Table~\ref{tab:settings}.
We can observe that under all settings, by pursuing trend awareness, the model performance is further improved.

\subsection{Empirical Study}
We investigate the influence of hyper-parameters (Eq.~\ref{eq:overall}) on BP4D. As in Fig.~\ref{fig:hyper}, by setting each $\lambda$ to the given seven values, we can observe that for all kinds of awareness, the performance is not very sensitive to the corresponding trade-off hyper-parameter. When $\lambda_{\rm rank}=0.1$, $\lambda_{\rm spd}=0.05$ and $\lambda_{\rm sub}=0.005$, the model achieves better performance.

\begin{figure}[!ht]
    \centering
    \includegraphics[width=0.42\textwidth]{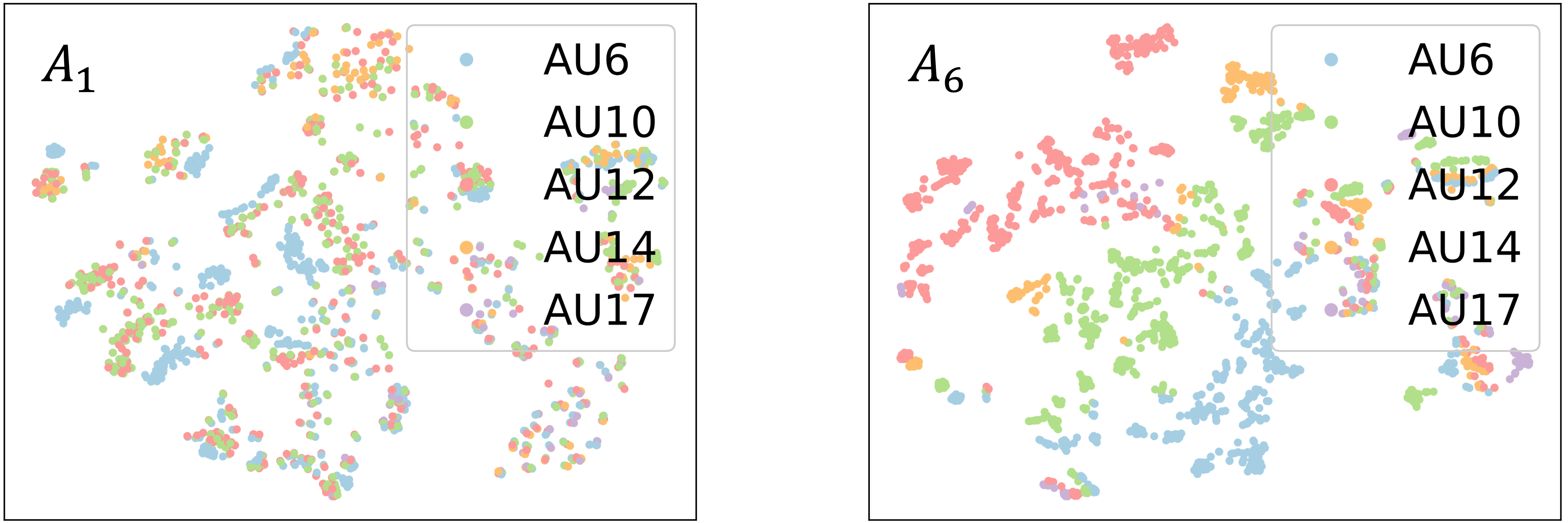}
    \caption{t-SNE visualization for AU features of one subject. Left: AU features from $A_1$. Right: AU features from $A_6$. 
    }
    \label{fig:tsne_au}
\end{figure}

\subsection{Qualitative Results}
Fig.~\ref{fig:ablation_case} shows that by pursuing ranking awareness, there are fewer jitters in the results of Model $A_2$. By pursuing speed awareness, the changing trend of intensity estimated by Model $A_5$ is more consistent with the corresponding facial appearance changes. With inter-trend subject awareness, for AUs with lower occurrence frequency such as AU14 and AU17, on which the model tends to overfit, some overall offsets on the changing trend of the estimated intensity are rectified by Model $A_6$, as in the right two tuples. And compared to FACS annotations, the estimated results are more accurate in some cases. \Eg, in the top-left tuple, there is no obvious difference between the last two frames, and thus the estimated intensity of AU6 barely changed. And in the bottom-left tuple, the woman pulls her lip corner a little bit with her dimple getting deeper in the marked frames, corresponding to a slight increase in the intensity of AU12.

Fig.~\ref{fig:tsne_sub} shows that AU features with different intensity labels extracted by Model $A_4$ (only with extra $\mathcal{L}_{\rm sub}$) are more compact, and AU features from one subject but with different intensity are far away from each other, which indicates that the learned AU-specific features are consistent among subjects.
Fig.~\ref{fig:tsne_au} shows that AU features of Model $A_6$ are more disentangled with each other than those of Model $A_1$, which demonstrates the superiority of trend-aware supervision in learning invariant AU features.

\section{Conclusion}
In this paper, we inspect the keyframe-based semi-supervised AU intensity estimation and identify the spurious correlation problem as the main challenge for achieving intensity estimation invariance. To this end, we propose trend-aware supervision to raise trend awareness during training. Extensive experiments show that all kinds of awareness are essential and help in learning invariant AU-specific features.

\bibliography{main}

\end{document}